\documentclass{article}

\usepackage{arxiv}

\usepackage[utf8]{inputenc} % allow utf-8 input
\usepackage[T1]{fontenc}    % use 8-bit T1 fonts
\usepackage{hyperref}       % hyperlinks
\usepackage{url}            % simple URL typesetting
\usepackage{booktabs}       % professional-quality tables
\usepackage{amsfonts}       % blackboard math symbols
\usepackage{nicefrac}       % compact symbols for 1/2, etc.
\usepackage{microtype}      % microtypography
\usepackage{lipsum}		% Can be removed after putting your text content
\usepackage{graphicx}
\usepackage{natbib}
\usepackage{doi}
\usepackage{pgfplots}
\usepackage{subcaption}
\usepackage[utf8]{inputenc}
\usepackage{float}
\usepackage{wrapfig}
\usepackage{booktabs}
\usepackage{multicol}
\usepackage{multirow}
\usepackage{caption} 
\usepackage{amsmath}
\usepackage{relsize}

\pgfplotsset{width=10cm,compat=1.9}

\usepgfplotslibrary{external}
\definecolor{light-gray}{gray}{0.95} 
\newcommand{\code}[1]{\colorbox{light-gray}{\texttt{#1}}}

\title{Learning Long Sequences in Spiking Neural Networks}

\author{ \href{https://orcid.org/0000-0003-2726-1860}{\includegraphics[scale=0.06]{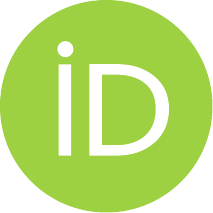}\hspace{1mm}Matei-Ioan Stan} \\
	Department of Computer Science\\
	The University of Manchester\\
	Manchester, United Kingdom \\
	\texttt{matei.stan@manchester.ac.uk} \\
	\And
	\href{https://orcid.org/0000-0003-1728-2828}{\includegraphics[scale=0.06]{orcid.pdf}\hspace{1mm} Oliver Rhodes} \\
        Department of Computer Science\\
	The University of Manchester\\
	Manchester, United Kingdom  \\
	\texttt{oliver.rhodes@manchester.ac.uk} \\
}

\renewcommand{\headeright}{}
\renewcommand{\undertitle}{}
\renewcommand{\shorttitle}{Learning Long Sequences in Spiking Neural Networks}

\hypersetup{
pdftitle={Learning Long Sequences in Spiking Neural Networks},
pdfsubject={q-bio.NC, q-bio.QM},
pdfauthor={Matei-Ioan Stan},
pdfkeywords={Spiking Neural Networks, State Space Models, Sequence Modelling, Long Range Dependencies},
}

\begin{document}
\maketitle

\begin{abstract}
     Spiking neural networks (SNNs) take inspiration from the brain to enable energy-efficient computations. Since the advent of Transformers, SNNs have struggled to compete with artificial networks on modern sequential tasks, as they inherit limitations from recurrent neural networks (RNNs), with the added challenge of training with non-differentiable binary spiking activations. However, a recent renewed interest in efficient alternatives to Transformers has given rise to state-of-the-art recurrent architectures named state space models (SSMs). This work systematically investigates, for the first time, the intersection of state-of-the-art SSMs with SNNs for long-range sequence modelling. Results suggest that SSM-based SNNs can outperform the Transformer on all tasks of a well-established long-range sequence modelling benchmark. It is also shown that SSM-based SNNs can outperform current state-of-the-art SNNs with fewer parameters on sequential image classification. Finally, a novel feature mixing layer is introduced, improving SNN accuracy while challenging assumptions about the role of binary activations in SNNs. This work paves the way for deploying powerful SSM-based architectures, such as large language models, to neuromorphic hardware for energy-efficient long-range sequence modelling. 
\end{abstract}

% keywords can be removed

\section{Introduction} \label{sec:intro}

Modelling long-range sequences is a fundamental component in solving many real-world challenges, with aplications ranging from processing biosignals such as electroencephalograms spanning tens of thousands of time steps \citep{tang2023modeling}, to comprehending and potentially writing large documents (e.g., novels, scientific papers) using large language models \citep{zhou2023recurrentgpt, liu2023lost}.  

\begin{figure}[H]
    \centering
        \begin{subfigure}[b]{0.467\textwidth}
            \centering        
            \includegraphics[width=\textwidth]{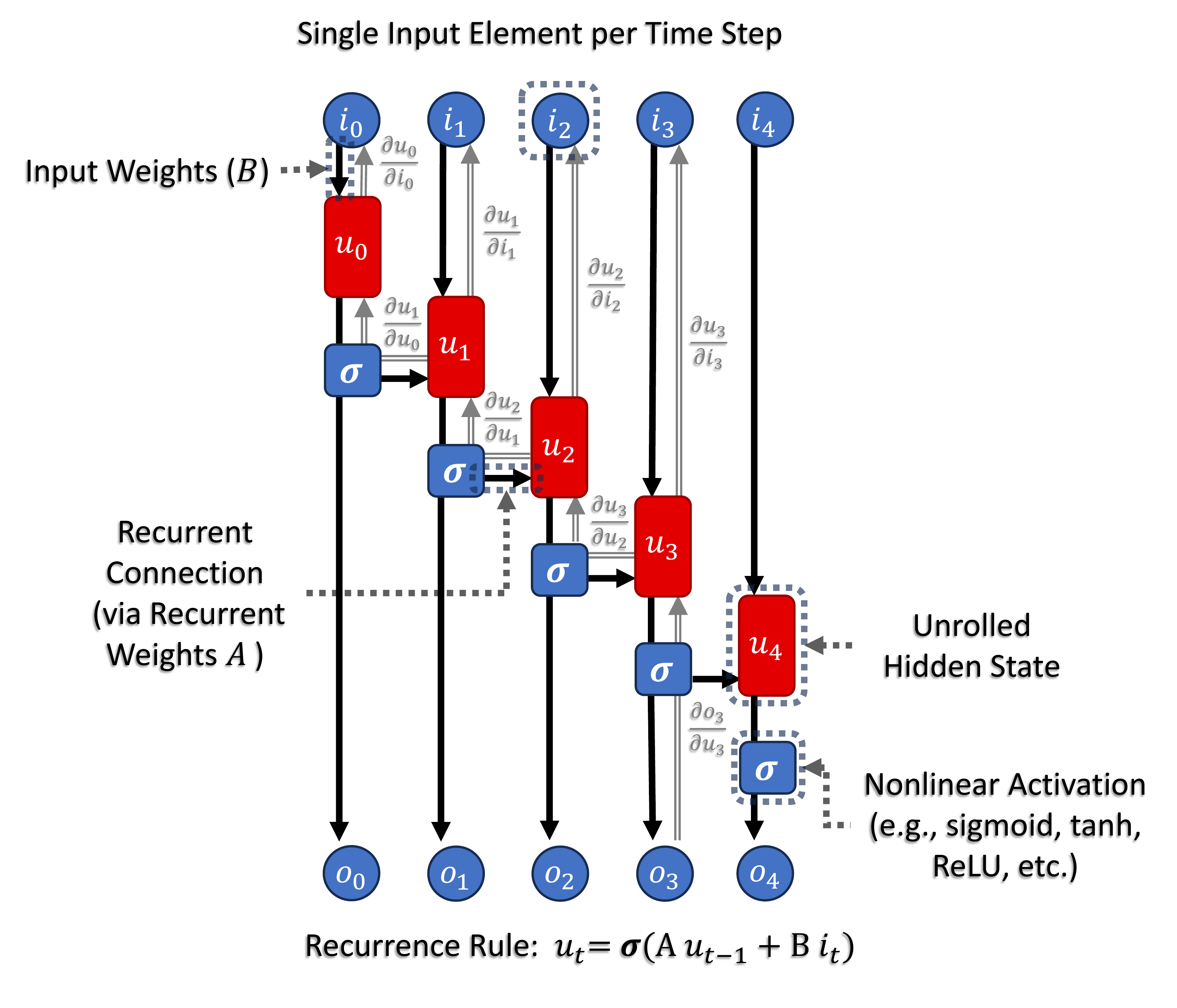}
            \caption{Unrolled Computations in Basic RNNs}
            \label{fig:unroll_rnn}
         \end{subfigure}
        \begin{subfigure}[b]{0.467\textwidth}
            \centering        
            \includegraphics[width=\textwidth]{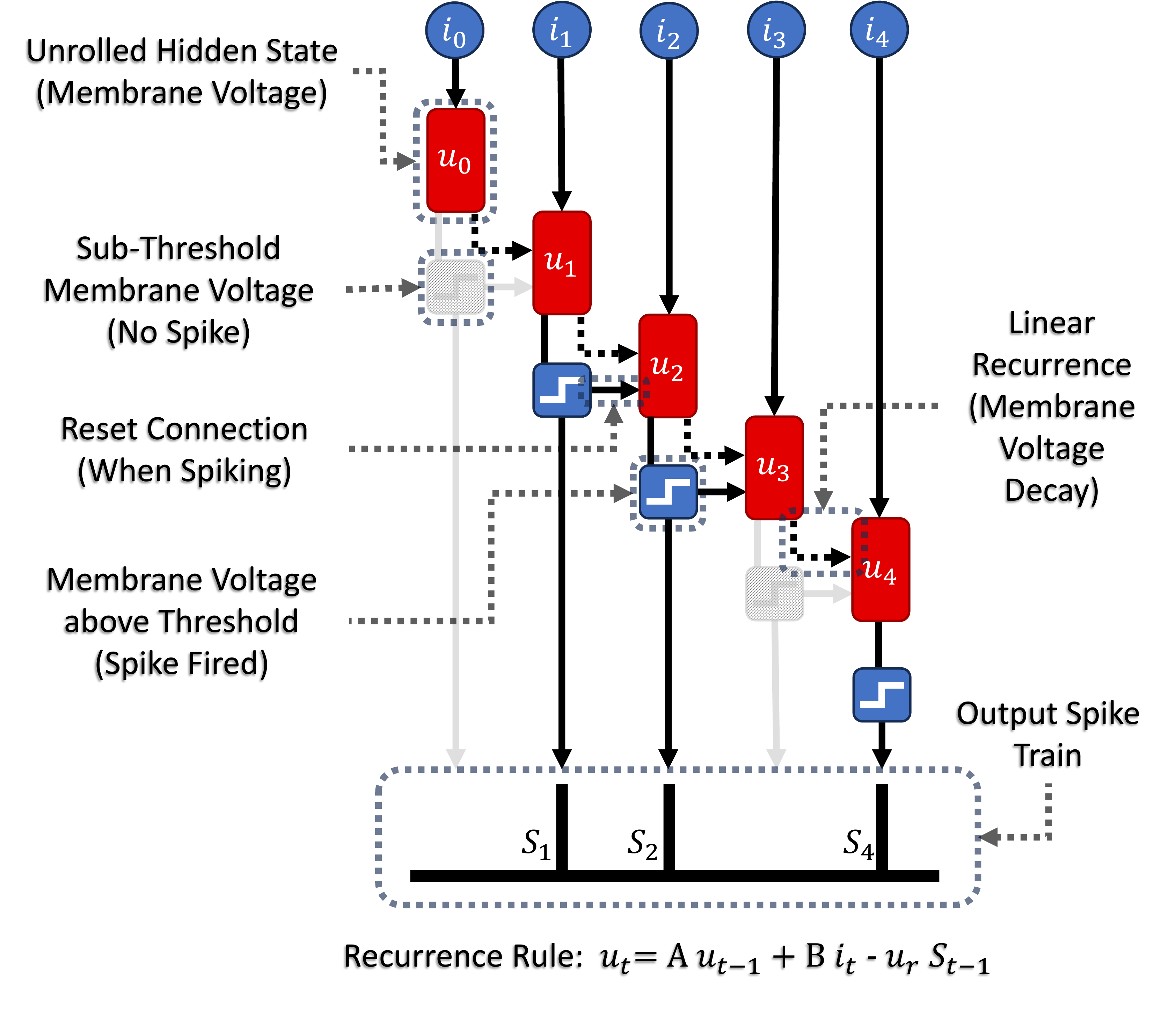}
            \caption{Unrolled Computations in SNNs}
            \label{fig:unroll_snn}
        \end{subfigure}\\
        \begin{subfigure}[b]{0.457\textwidth}
            \centering        
            \includegraphics[width=\textwidth]{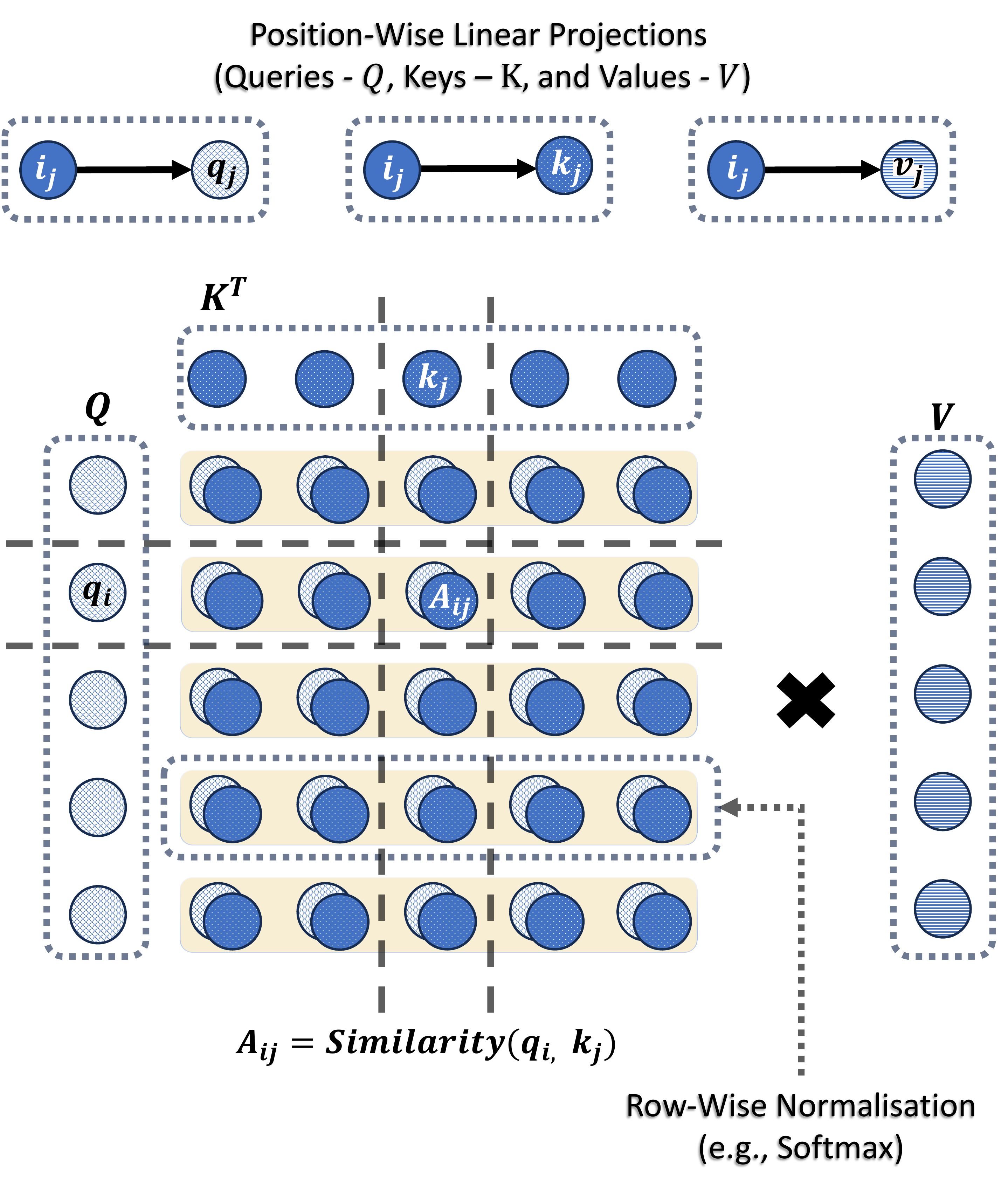}
            \caption{Self-Attention}
            \label{fig:self_attention}
        \end{subfigure}
        \begin{subfigure}[b]{0.457\textwidth}
            \centering        
            \includegraphics[width=\textwidth]{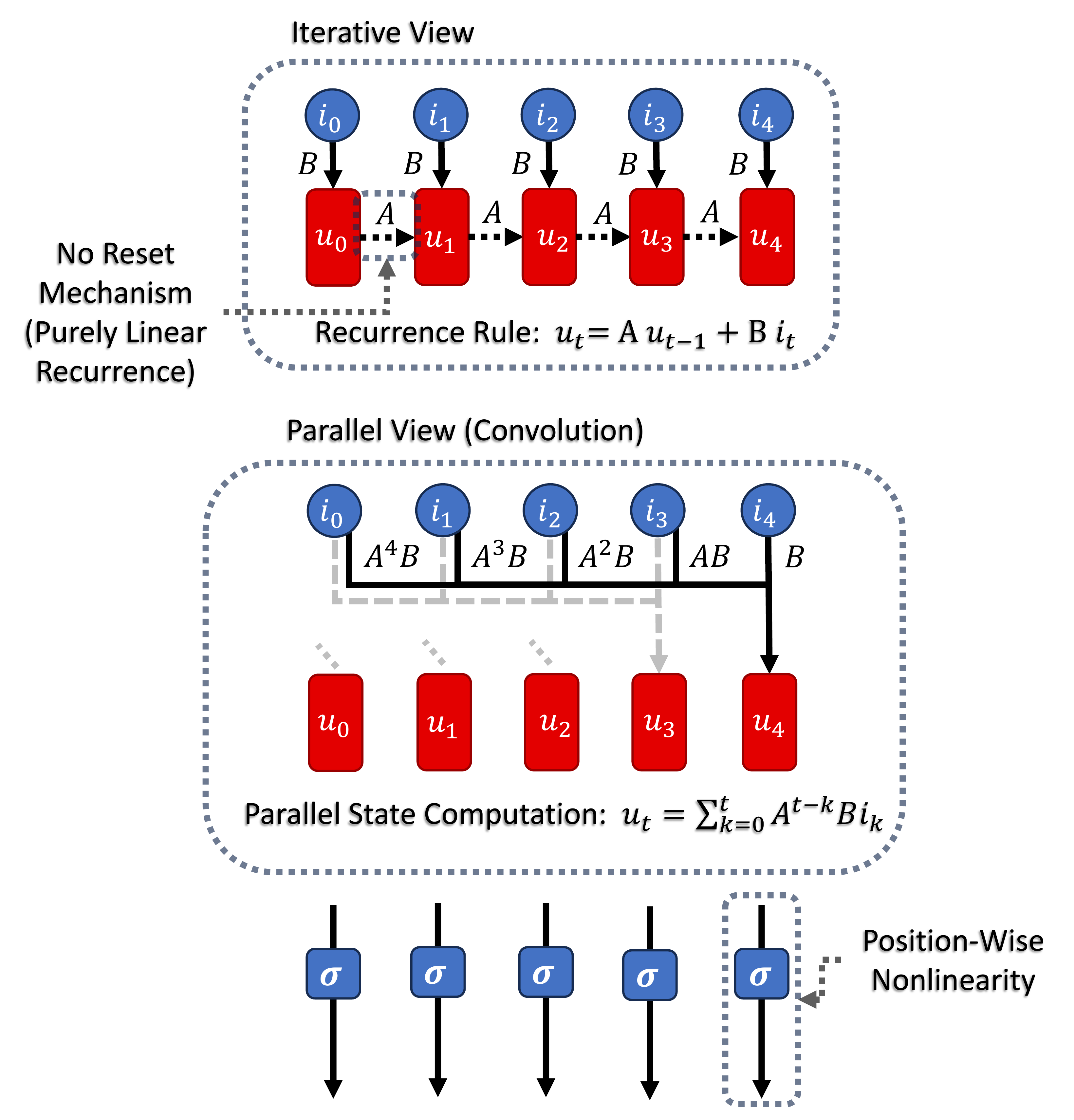}
            \caption{Unrolled Computations in SSMs}
            \label{fig:unrolled_ssm}
        \end{subfigure}\\
        
     \hfill
        \caption{\textbf{Example Computational Graphs for Sequence Models}. Subfigure~\ref{fig:unroll_rnn}, shows how basic RNNs perform computations over time. Of note is the inclusion of nonlinearities between time steps, which entail iterative computations. In addition, one can observe how, during the backward pass using BPTT, credit assignment between time steps $\frac{\partial h_{p}}{\partial h_{q}}$, where $q << p$, involves numerous repeated multiplications which can cause vanishing or exploding gradients. Subfigure~\ref{fig:unroll_snn}, highlights the structural similarities between SNNs and RNNs. One important difference stems from the addition of a linear recurrence based on leaky membrane voltages in neurons such as Leaky Integrate-and-Fire neurons in SNNs. Moreover, the defining feature of SNNs is the neuron outputs consisting of sparse binary spike trains. Subfigure~\ref{fig:self_attention}, underlines the parallel nature of Transformers, where input history is no longer compressed within an evolving network state. The attention matrix containing all pair-wise similarities between tokens in the input sequence is multiplied with the $V$ projection of the inputs in dense and large-scale matrix-matrix multiplication, which is unfavourable for neuromorphic hardware implementation. Subfigure~\ref{fig:unrolled_ssm}, illustrates the dual interpretation of recurrences in linear time-invariant SSMs. In architectures such as S4 \citet{gu2021efficiently}, individual SSM units are single-input single-output (SISO). The scalar input ($i_t$) is projected onto high-dimensional space using $B \in \mathbb{R}^{d}$ at each time step. The state of the model ($u_t$) evolves over time using the transition matrix $A \in \mathbb{R}^{d \times d}$. SSM-based neural networks use the initialisation of $A$ and $B$ to implicitly encode projections of input signals onto an orthogonal polynomial basis. To produce a scalar output ($y_t$), the state vector is linearly projected back onto a single dimension using a vector $C \in \mathbb{R}^{d}$. }
        \label{fig:recurrence}
\end{figure}

Deep learning methods have established themselves as state-of-the-art solutions for numerous challenging tasks, including learning functions defined over variable-length input sequences. Recurrent neural network (RNN) architectures emerged early on as strong contenders for this purpose. They compress sequences by incorporating input elements one at a time, using only $\mathcal{O}(1)$ operations with respect to the sequence length to process each input token and sharing parameters between time steps (Figure \ref{fig:unroll_rnn}). Notably, RNNs are partially inspired by cognitive and neurological computational principles \citep{lipton2015critical}. Hence, perhaps unsurprisingly, they also underpin another class of biologically grounded architectures - spiking neural networks (SNNs) (Figure \ref{fig:unroll_snn}). SNNs process sequences using simplified mathematical models of biological neurons that relay internal computations using sparse patterns of binary spikes \citep{maass1997networks}. The aim is to emulate the brain's efficient neural coding, which enables computing with a fraction of the energy required by modern von Neumann machines \citep{hasler2017special}.

RNNs are affected by vanishing and exploding gradients \citep{pascanu2013difficulty}, stemming from unstable recurrent weight initialisation and the use of backpropagation through time (BPTT) (Figure~\ref{fig:unroll_rnn}). These phenomena hinder learning long-range dependencies in RNNs, and while they can be mitigated to some extent by gating mechanisms such as long short-term memory (LSTM)  \citep{hochreiter1997long}, they difficult to eliminate entirely. In addition, traditional RNNs apply nonlinearities at each time step ($\sigma$ in Figure  \ref{fig:unroll_rnn}), which requires iterative computations. This approach is non-problematic at inference, where input sequence elements are unknown ahead of time. However, RNN forward passes become prohibitively slow at training time for long sequences, since they cannot take advantage of GPU parallelisation, owing to the nonlinear state propagation \citep{10191884, orvieto2023resurrecting, kalchbrenner2016neural}.  

Additional challenges arise in SNN learning, as binary spiking is non-differentiable, which prohibits training SNNs directly with backpropagation. One solution is to train an artificial neural network (ANN) and then convert its continuous activations to spikes \citep{diehl2015fast}. However, this approach introduces additional latency during inference and is often prone to excessive firing, which can damage the energy efficiency of the network \citep{davidson2021comparison}. Another solution is to train SNNs directly using surrogate gradients in the backward pass \citep{neftci2019surrogate}. Nevertheless, even with surrogate-based training, SNNs are still generally outperformed by ANNs such as LSTMs \citep{malcom2023comprehensive}. 

The RNN limitations mentioned above are overcome by the Transformer \citep{vaswani2017attention}, which directly compresses the context for each token by measuring its relationship to all other elements (Figure \ref{fig:self_attention}). Besides improving performance, the Transformer's core component, self-attention, can be easily parallelised through GPU-friendly matrix multiplication, which accelerates training relative to RNNs \citep{zeyer2019comparison}. Consequently, Transformer blocks have been crucial in establishing the current golden age of ever-larger pre-trained models \citep{min2023recent}. 

The parallel and dense matrix multiplications that have entrenched the Transformer as arguably the de facto standard in sequence modelling also accentuated the structural differences between SNNs and ANNs. SNNs are built for deployment on neuromorphic computing platforms such as Intel Loihi  \citep{davies2021advancing}, which can potentially enable orders of magnitude lower energy consumption compared to traditional computers. These efficiencies are partly supported by representing information as sparse events identified by their address. Spike events then "excite" the targeted synapses asynchronously, with accumulation occurring within the postsynaptic neurons' internal states. This enables addition-based feature mixing, reducing costly Multiply-and-Accumulate (MAC) operations \citep{li2023brain}. Massive parallel matrix multiplications, as self-attention requires, can be seen as antagonistic to this event-driven and brain-inspired computing philosophy. Therefore, lessons from Transformer-based research have seen relatively limited adoption in SNNs by comparison \citep{zhou2022spikformer, zhu2023spikegpt, yao2023spike}. 

Nevertheless, self-attention suffers a quadratic computational cost with respect to sequence length \citep{tay2020efficient}, which effectively limits scaling to longer sequences. In addition, training and inference for large-scale Transformer-based models have seen significant increases in energy requirements, leading to considerable carbon emissions \citep{strubell2019energy}. This highlights the need for energy-efficient models which scale better with input length, a role recurrent SNNs are potentially well-positioned to fill.

The quadratic computational cost has motivated a recent resurgence in RNN research interest. Receptance Weighted Key Value (RWKV) \citep{peng2023rwkv}, exemplifies research focused on reducing the computational complexity of Transformers.  It is essentially a recurrent self-attention adaptation allowing $\mathcal{O}(1)$ iterative deployment. Another area of research is focused on deriving RNNs with theoretical guarantees regarding long-range modelling properties. For example, the Legendre Memory Unit (LMU), takes inspiration from hippocampal neurons to augment RNN nonlinear state propagation with a linear memory component \citet{voelker2019legendre}. The memory unit is constructed using linear projections of input signals onto a Legendre orthogonal polynomial basis. The result is a multidimensional cell state that is theoretically guaranteed to encode a sliding window of a given number of past inputs. This enabled the LMU to become the first recurrent model to successfully capture temporal dependencies on the scale of 100,000 time steps \citep{voelker2019legendre}. \citet{chilkuri2021parallelizing} remove the remaining nonlinear recurrences in the LMU to obtain a linear time-invariant (LTI) structure with position-wise activations (Figure~\ref{fig:unrolled_ssm}). LTI systems have the property of having two equivalent formulations: iterative propagation of the system's state by repeated application of the linear recurrence; or a convolution of the input signal with a global filter implicitly parametrised by the linear recurrence parameters. Crucially, convolutions can be implemented efficiently using subquadratic $\mathcal{O}(Nlogn(N))$ fast Fourier transforms (FFTs) \citep{gu2021combining}. In sum, linear RNNs have the desirable property of GPU-friendly parallelisability at training time while retaining efficient iterative deployment for inference. 

Structured state space models (S4), introduced by \citet{gu2021efficiently}, generalise the parallelisable LMU by exploring intialisation of recurrent weights based on alternative orthogonal polynomial bases \citep{gu2020hippo}. This enables input history compression with biases different from sliding windows (e.g., exponentially decaying). S4 established the state space model (SSM) class of neural architectures as state-of-the-art methods on several long-range sequence modelling tasks. For example, it outperformed the Transformer in terms of accuracy by an average of 30\% on the tasks of the challenging \textbf{Long Range Arena} (\textbf{LRA}) benchmark \citep{tay2020long}. Nevertheless, as highlighted by Figure \ref{fig:unrolled_ssm}), computing the kernel for the global convolutions entails raising the transition matrix  ($A \in \mathbb{R}^{d \times d}$) to high powers, which can become slow for large values of $d$. \citet{gu2021efficiently} overcome this using efficient multiplication of low-rank approximations of $A$. Subsequent works have further simplified this process by establishing almost equally effective diagonal initialisation schemes for $A \in \mathbb{C}^{d}$ \citep{orvieto2023resurrecting, gu2022parameterization, gupta2022diagonal}. Diagonal transition matrices are also better suited for deployment to neuromorphic hardware since they allow for iterative state propagation based on Hadamard products rather than dense vector-matrix multiplication. This also entails state variables evolving independently over time, with an implementation reminiscent of exponentially decaying synapses and membrane voltages in spiking neurons, topics present in neuromorphic hardware research \citep{eissa2021hardware}. 

\textbf{Related Work} The renewed interest in RNNs has also inspired works applying these new techniques to SNNs. Some investigate stacking state-of-the-art ANN layers and well-studied neuromorphic Leaky Integrate-and-Fire (LIF) neurons (Equation \ref{eq:lif}) -- a leading example of this line of research being SpikeGPT \citep{zhu2023spikegpt}. The authors present the largest SNN language model to date, constructed by feeding outputs from RWKV layers into LIF neurons, which enable sparse spike-based feature mixing. While the RWKV layers could be parallelised as convolutions, the inclusion of LIF neurons imposes iterative computations during training, as highlighted by Figure~\ref{fig:unroll_snn}. Another example of this research direction is SpikeS4 \citep{du2023spiking}, where LIF neurons are stacked onto S4 layers. 

Other works have focused on parallelising the LIF neuron itself. For instance, \citet{fang2023parallel} present leaky integration strategies based either on multiplying the entire length-$N$ input sequences by $N \times N$ positional encoding matrices in a similar fashion to self-attention matrix multiplication (Figure~\ref{fig:self_attention}) or linearly integrating over an explicit buffer containing sliding windows of the input. Binary spiking is then applied in a position-wise manner. Crucially, neither strategy is formulated for $\mathcal{O}(1)$ iterative deployment. \citet{10191884} bring SNNs closer to SSMs by exploring both iterative and parallel computations for linear recurrences. However, compared to the linear memory unit of the LMU and other SSMs, both \citet{fang2023parallel} and \citet{10191884} focus on neurons with scalar internal states. High-dimensional internal states in SSMs constitute linear relationships between input tokens. For single-input single-output (SISO) SSMs with a $d$-dimensional internal state ($u$) as used in S4, the same linear relationships can be computed through a convolution of the scalar input signal with a scalar global kernel (Section~\ref{sec:ssm}). Crucially, this means the $d$-dimensional states $u$ do not have to be explicitly stored during training, reducing memory requirements \citep{gu2021efficiently}. If, instead, a nonlinearity is applied position-wise to each of the $d$ dimensions of the state, then they have to be explicitly materialised. The spiking neurons with scalar states in \citet{10191884} and \citet{fang2023parallel} manifest this structural pitfall, which may prohibit scaling these methods for challenging long-range tasks or using them to build large pre-trained architectures \citep{gu2021combining}. Moreover, the initialisation of the decay factor in \citet{10191884} and \citet{fang2023parallel} is constant between all neurons, which hinders learning dependencies across varying time scales \citep{orvieto2023resurrecting, hermans2010memory}. 

Hence, one can notice a significant gap in research at the intersection of state-of-the-art SSMs and SNNs. To the authors' knowledge, SNNs that borrow powerful initialisation and parameterisation techniques from SSM architectures, such as S4, while retaining their parallelisability, have not been studied so far. Consequently, this paper employs SSM-based SNNs to investigate whether SNNs can eventually become viable energy-efficient alternatives to state-of-the-art ANNs for challenging long-range sequence modelling tasks. Two core questions are addressed in this regard: \textbf{(a)} Do binary spiking activations inherently prevent SNNs from competing with ANNs on long-range sequence modelling? \textbf{(b)} In case they fundamentally hinder performance, should binary spikes necessarily define SNNs? 

\begin{figure}[H]
	\centering
        \includegraphics[width=0.9\textwidth]{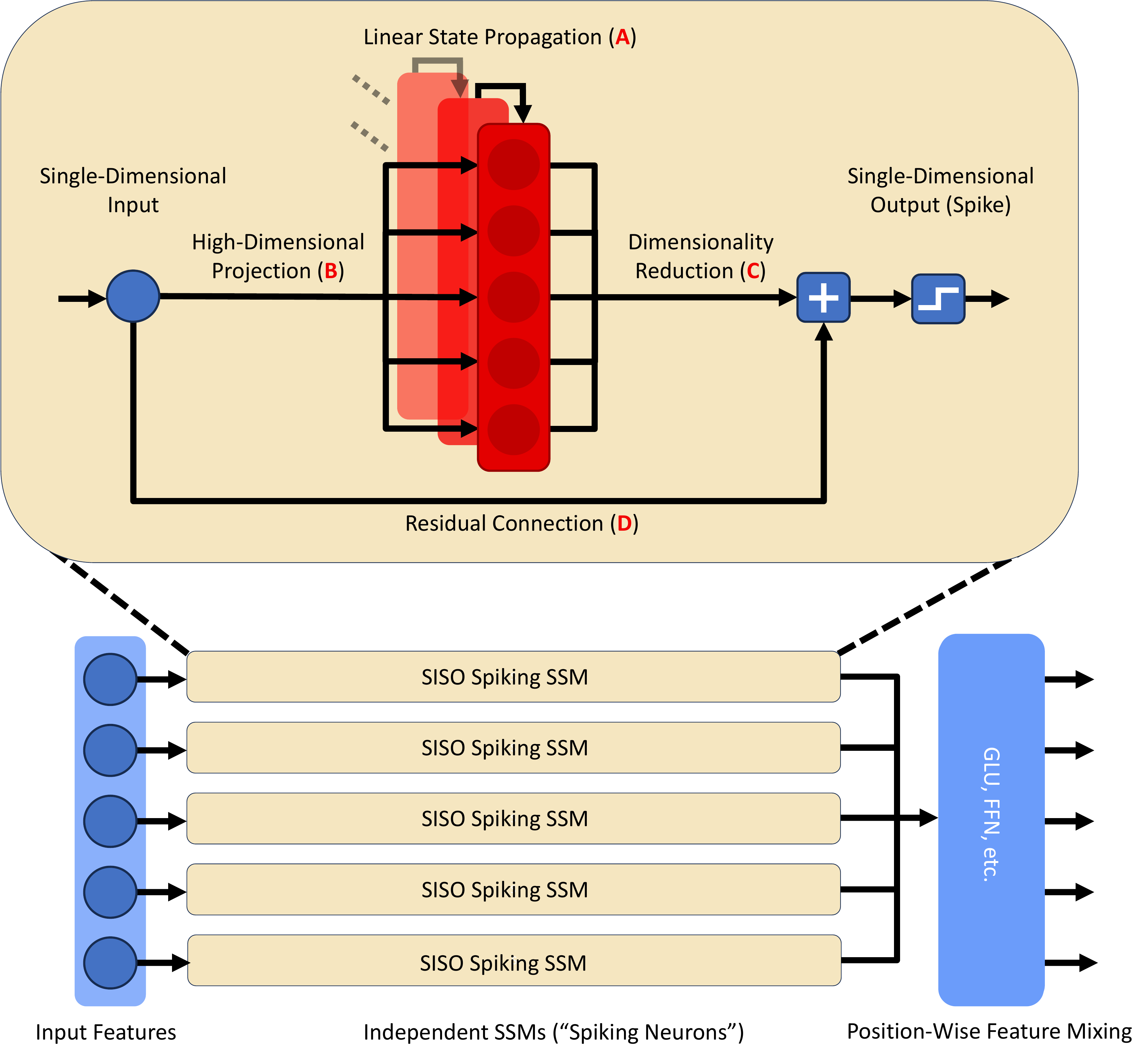}
	\caption{\textbf{Binary SSM Layer} At each time step, a Binary SSM layer consists of independent single-input single-output (SISO) SSM "neurons".  Binary activations are applied element-wise per each SSM output before position-wise feature mixing to avoid dense vector-matrix multiplication. }
	\label{fig:binary_s4d}
\end{figure}

To answer \textbf{(a)}, this paper formulates \textbf{Binary SSMs} as SSM-based SNNs (Figure~\ref{fig:binary_s4d}). The models are implemented using S4D initialisation \citep{gu2022parameterization}, and the performance of the resulting \textbf{Binary S4D} (Section~\ref{sec:binary_s4d}) is evaluated. Answering \textbf{(b)} requires challenging the role of binary activations in SNNs, which is mainly to avoid MAC operations for feature mixing. Conversely, this approach assumes that mixing continuous features is synonymous with relying on MAC operations. The \textbf{Gated Spiking Unit} (GSU) is formulated here for the first time in order to challenge this assumption (see Section \ref{sec:gsu} for further details). The GSU is a position-wise feature mixing layer inspired by the Gated Linear Unit (GLU) \citep{dauphin2016language} based on two parallel streams. Continuous SSM features $\in \mathbb{R}$ are mixed using ternary weights $\in \{-1, 0, 1\}$ \citep{zhu2016trained}, while ternarised SSM outputs are mixed using a continuous-valued linear layer. The final output of the GSU is the Hadamard product of the feature vectors resulting from the two streams. Both streams require only inexpensive additions/subtractions, avoiding MAC operations. Most importantly, as opposed to binarisation in traditional SNNs, the GSU avoids vanishing gradients by allowing backpropagation through non-saturating activations \citep{gulcehre2016noisy}. GLU-inspired layers in SNNs have been studied before, but only in the context of mixing binary spike features with continuous weights \citep{zhu2023spikegpt}.

The remainder of the paper is structured as follows. First, Binary SSMs (Section~\ref{sec:binary_s4d}) are compared to the GSU (Section~\ref{sec:gsu}) and baseline state-of-the-art ANN models on the \textbf{Long Range Arena} benchmark \citep{tay2020long} (\textbf{LRA}). This is the first time SNNs are comprehensively and systematically studied on significantly longer sequences than standard neuromorphic datasets. Second, Binary SSMs and the GSU are compared to, and shown to outperform, current state-of-the-art SNNs on \textbf{sequential MNIST} (\textbf{sMNIST}) classification \citep{le2015simple}, under similar constraints (Section \ref{sec:smnist_config}). Third, the effect of the surrogate gradient function on classification accuracy is highlighted for \textbf{sequential CIFAR10} (\textbf{sCIFAR10}). Finally, the most difficult long-range modelling task in the LRA, \textbf{Path-X}, is used to compare binary activations and the GSU with continuous-valued saturating activation functions (arctan, fast sigmoid). This highlights that non-differentiable binary activations are upper-bounded in accuracy by continuous-valued saturating activations, which themselves lag far behind non-saturating activations in deep SSM models. 

\section{Results}

\begin{figure}[H]
        \begin{subfigure}[b]{0.25\textwidth}
            \centering
            \includegraphics[scale=0.575]{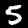}
            \includegraphics[scale=0.575]{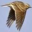}
            \includegraphics[scale=0.575]{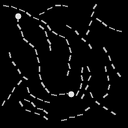}
            \caption{Image Samples (To Scale)}
            \label{fig:original_samples}
        \end{subfigure}
        \begin{subfigure}[b]{0.155\textwidth}
            \centering        
            \includegraphics[width=\textwidth]{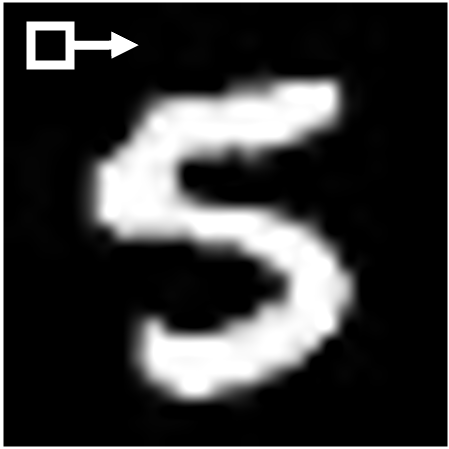}
            \caption{Image Flattening}
            \label{fig:smnist_bellec}
        \end{subfigure}
        \begin{subfigure}[b]{0.57\textwidth}
            \centering        
            \includegraphics[width=\textwidth]{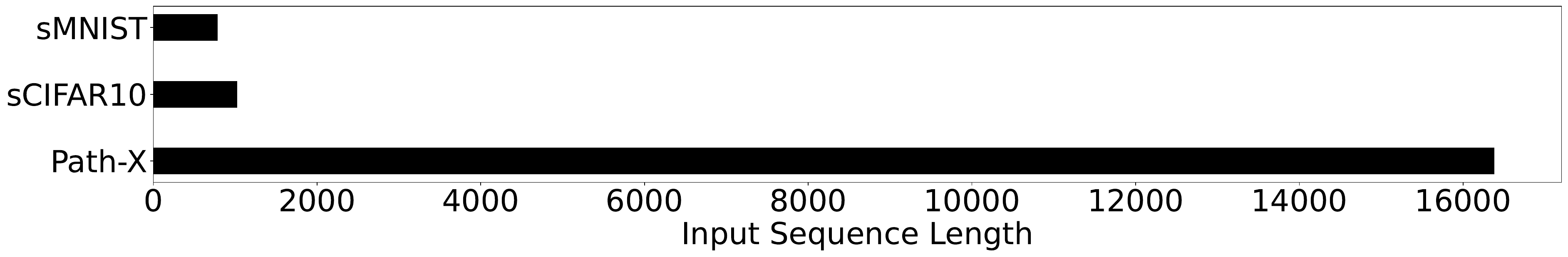}
            \caption{Flattened Image Sample Length}
            \label{fig:flattened_samples}
         \end{subfigure}
         \hfill
            \caption{\textbf{Input Scales} Subfigure \ref{fig:original_samples} shows relative sizes of samples from (left to right) \textbf{MNIST}, \textbf{CIFAR10} and \textbf{Path-X}, with respective resolutions of 28x28 (784), 32x32 (1024), and 128x128 (16384). Subfigure \ref{fig:smnist_bellec} shows the flattening process used in all image-based sequential tasks (adapted from \citet{bellec2018long}). Subfigure \ref{fig:flattened_samples} visualises how the lengths of the flattened image samples compare. One can easily observe from \ref{fig:original_samples} and \ref{fig:flattened_samples} that \textbf{Path-X} contains input sequences more than twenty times longer than \textbf{sequential MNIST}, commonly used for probing SNN long-range dependencies. }
            \label{fig:samples}
\end{figure}

The selection of evaluation tasks is guided by the need to compare the proposed architectures with state-of-the-art in both neuromorphic and broader sequence modelling research. The neuromorphic community has widely embraced variants of the MNIST dataset as standard benchmarks \citep{malcom2023comprehensive}, therefore its sequential variant (\textbf{sMNIST}) is employed here. \textbf{sMNIST} consists of flattening the 28x28 MNIST samples to 784-long sequences by appending one pixel at a time to a scalar list, row-by-row (Figure \ref{fig:smnist_bellec}). 

State-of-the-art sequence modelling architectures are typically evaluated using the \textbf{Long Range Arena} (\textbf{LRA}) \citep{tay2020long}, consisting of a suite of six tasks. Long \textbf{ListOps}, first introduced by \citet{nangia2018listops}, requires capturing latent hierarchies by parsing nested operations, forming sequences of 2k elements. The \textbf{Text} task is built around the binary classification of byte-level (character-level) IMDB reviews, with sequence lengths fixed at 4k tokens. \textbf{Retrieval} measures how well models can compress byte-level document information to classify two documents' mutual similarity. Each sample consists of two concatenated documents totalling 8k input sequence elements. The \textbf{Image} task is constructed similarly to \textbf{sMNIST}, flattening out CIFAR10 images \citep{krizhevsky2009learning} into 1024-long sequences of gray-scale-valued pixels for classification (Figure~\ref{fig:flattened_samples}). Finally, \textbf{Pathfinder} and \textbf{Path-X} follow the same aforementioned flattening procedure for the binary classification of images in which two points are either connected or not by a dotted path (rightmost sample in Figure~\ref{fig:original_samples}). While the baseline \textbf{Pathfinder} task consists of images with resolutions of 32x32 (sequence length of 1024), \textbf{Path-X} employs samples with resolutions of 128x128, resulting in sequences of more than 16k elements (Figure~\ref{fig:flattened_samples}). 

\begin{figure}[H]
        \centering
        \includegraphics[width=0.9\textwidth]{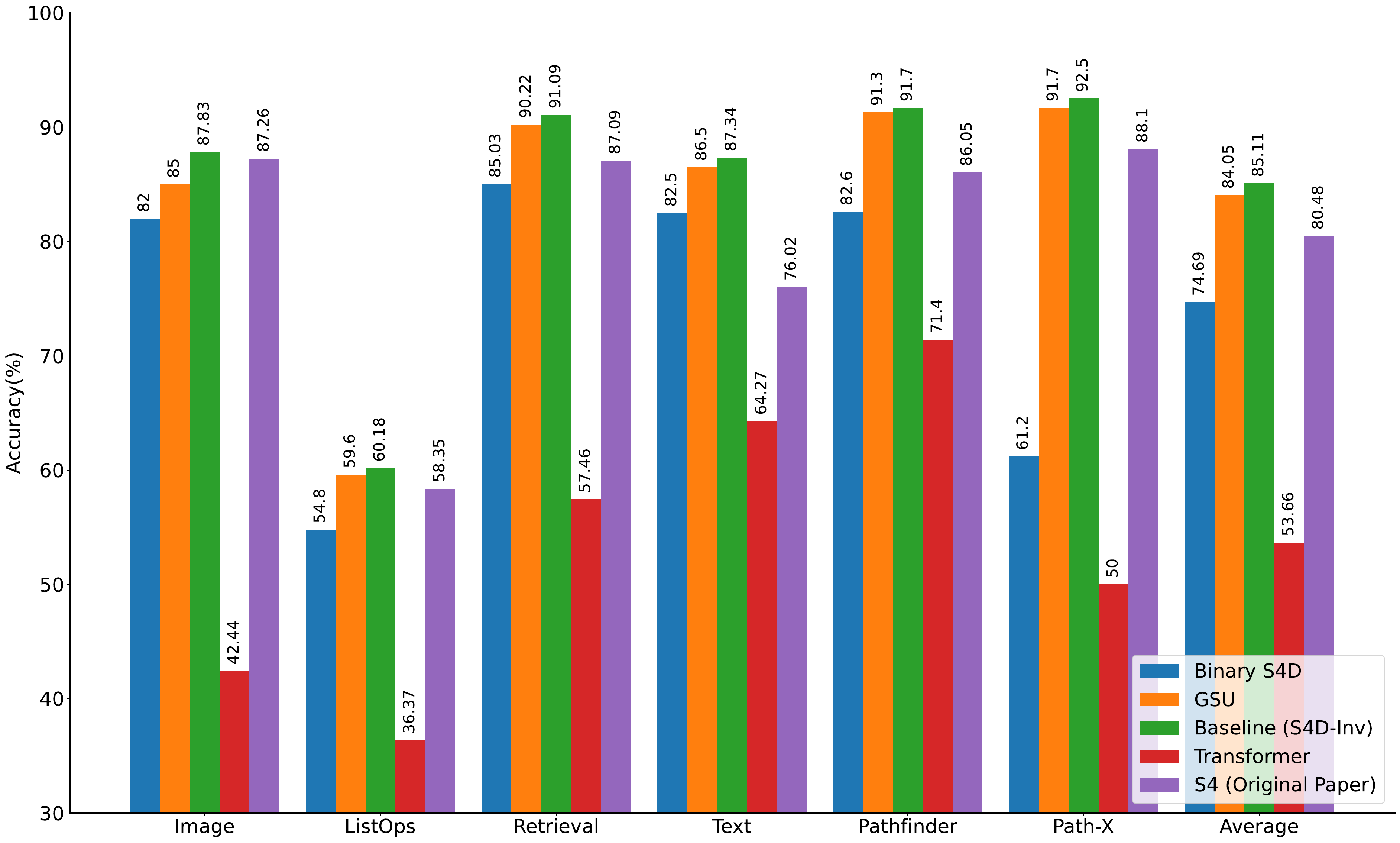}
        \caption{\textbf{Accuracy on the LRA benchmark.} Binary S4D performs on average more than 10\% worse than the baseline but still over 20\% better than the Transformer. The GSU achieves 1.06\% lower accuracy than the baseline on average, and at most just 2.83\% below the baseline on \textbf{Image}. On \textbf{Path-X}, Binary S4D has 30\% lower accuracy than the baseline yet still manages to outperform the Transformer by 11.2\%. }
        \label{fig:acc_lra}
\end{figure}

\subsection{\textbf{LRA Accuracy}} \label{sec:lra_acc}

The S4 model \citep{gu2021efficiently}, and subsequent variants \citep{gu2022parameterization}, have established themselves among state-of-the-art solutions on the \textbf{LRA} benchmark. For example, the original S4 model outperforms the Transformer on the \textbf{LRA} tasks by an average of more than 30\% in accuracy. Most importantly, on the longest and most challenging task, \textbf{Path-X}, S4 reaches 88.1\%, while the Transformer fails to converge beyond random accuracy (50\%). The baseline employed here, an S4D-Lin model (see Sections~\ref{sec:initialisation} and \ref{sec:lra_config}), achieves an even higher accuracy of 92.5\% on \textbf{Path-X}. 

The Binary S4D model (Section~\ref{sec:binary_s4d}), proposed in this paper, is trained and evaluated on all tasks within the LRA, to explore how binarisation impacts baseline performance. Figure~\ref{fig:acc_lra}, shows that Binary S4D lags behind the baseline in all tasks of the \textbf{LRA}. On the \textbf{Image}, \textbf{Listops}, \textbf{Retrieval}, and \textbf{Text} tasks, binary spikes impose at most a 6\% accuracy penalty. Accuracy is more strongly degraded on  \textbf{Pathfinder} and \textbf{Path-X}, where Binary S4D achieves 82.6\% and 61.2\%, respectively, compared to 91.7\% and 92.5\% for the baseline. Nonetheless, Binary S4D outperforms the Transformer on all tasks of the \textbf{LRA}, by an average margin of more than 20\%. Even where Binary S4D accuracy is significantly lower than the baseline, on \textbf{Path-X}, it still outperforms the Transformer by 11.2\%. As such, one can argue that SSM-based SNNs, such as Binary S4D, have stronger long-range modelling capabilities than basic Transformers, as indicated by the results on the \textbf{LRA}. Moreover, for \textbf{Path-X}, Transformer baseline accuracy is obtained with approximately 600k parameters \citep{tay2020long, gu2021efficiently}, while Binary S4D uses less than 200k. Since Binary S4D outperforms the Transformer on all tasks of the \textbf{LRA}, one can reasonably argue that binary spiking does not inherently prevent SNNs from exhibiting competitive performance with respect to other ANN architectures. This result contributes to answering question \textbf{(a)} posed in Section~\ref{sec:intro}. 

The GSU (Section~\ref{sec:gsu}), is also trained and evaluated on all tasks within the LRA, with performance presented in Figure~\ref{fig:acc_lra}, to highlight the improvements associated with non-saturating activations. The GSU model lags on average 1.06\% behind the baseline S4D-Inv model. While the largest discrepancy is on \textbf{Image}, it is still only 2.83\%. Interestingly, on \textbf{Path-X}, the GSU is only 0.8\% below the baseline (where Binary S4D dropped more than 30\%). Generally, it can be argued that the GSU achieves comparable accuracies to the baseline S4D-Inv model on all tasks of the \textbf{LRA}. Similarly, the GSU outperforms the Transformer on average by over 30\% on the \textbf{LRA}.

\begin{center}
    \begin{table}[H]
        \centering
        \begin{tabular}{lllll}
            \toprule
            
            Model                        & SSM Size & Parallelisable    & No. Trainable Parameters  & Accuracy \\
            
            \midrule
            
            \multirow{2}{*}{Binary S4D}  & 2        & Yes               & 68.9k                     & 99.1\%   \\
                                         & 64       & Yes               & 118k                      & 99.4\%   \\
                                         
            \midrule
            
            \multirow{2}{*}{GSU}         & 2        & Yes               & 37.9k                     & 99.2\%   \\
                                         & 64       & Yes               & 85.5k                     & 99.4\%   \\
            \midrule
            
            SRNN \citep{yin2021accurate} & N/A      & No                & 156k (estimate)           & 98.7\%   \\
            
            \midrule
            
            LSNN \citep{bellec2018long}  & N/A      & No                & 66k                       & 97.1\%   \\
            
            \bottomrule
            
        \end{tabular}
        \caption{\textbf{Accuracy on sMNIST} Binary S4D and GSU outperform current state-of-the-art SNNs, using fewer parameters. }
        \label{mnist_results_table}
    \end{table}
    
\end{center}

\subsection{\textbf{Sequential MNIST Accuracy}} \label{sec:smnist_acc}

The proposed Binary S4D and the GSU are compared to state-of-the-art SNNs using the \textbf{sMNIST} task. The landmark findings of \citet{bellec2018long} brought SNNs closer to matching LSTM accuracy on the well-documented \textbf{sMNIST} classification task. \citet{yin2021accurate} further built upon this result establishing current state-of-the-art SNN accuracy. The results in Table~\ref{mnist_results_table}, show Binary S4D models outperforming both methods, reaching an accuracy of 99.1\%, constituting state-of-the-art accuracy for SNNs to the best of the authors' knowledge. Expanding the state size for Binary S4D ($u \in \mathbb{C}^{64}$) further improves accuracy to 99.4\%. The GSU marginally improves accuracy in the $u \in \mathbb{C}^2$ configuration to 99.2\% and performs identically for the $u \in \mathbb{C}^{64}$ configuration. Both the GSU and Binary S4D models require fewer parameters to achieve higher accuracy than the SRNN \citep{yin2021accurate} while also allowing for parallelisable training. It is worth mentioning that while the SSM state size is two ($u \in \mathbb{C}^2$), the parameters of the two dimensions are conjugates, meaning the SSM requires training of only one set of parameters for both dimensions \citep{gu2022parameterization}. 

\begin{figure}[H]
        \centering
        \includegraphics[width=0.9\textwidth]{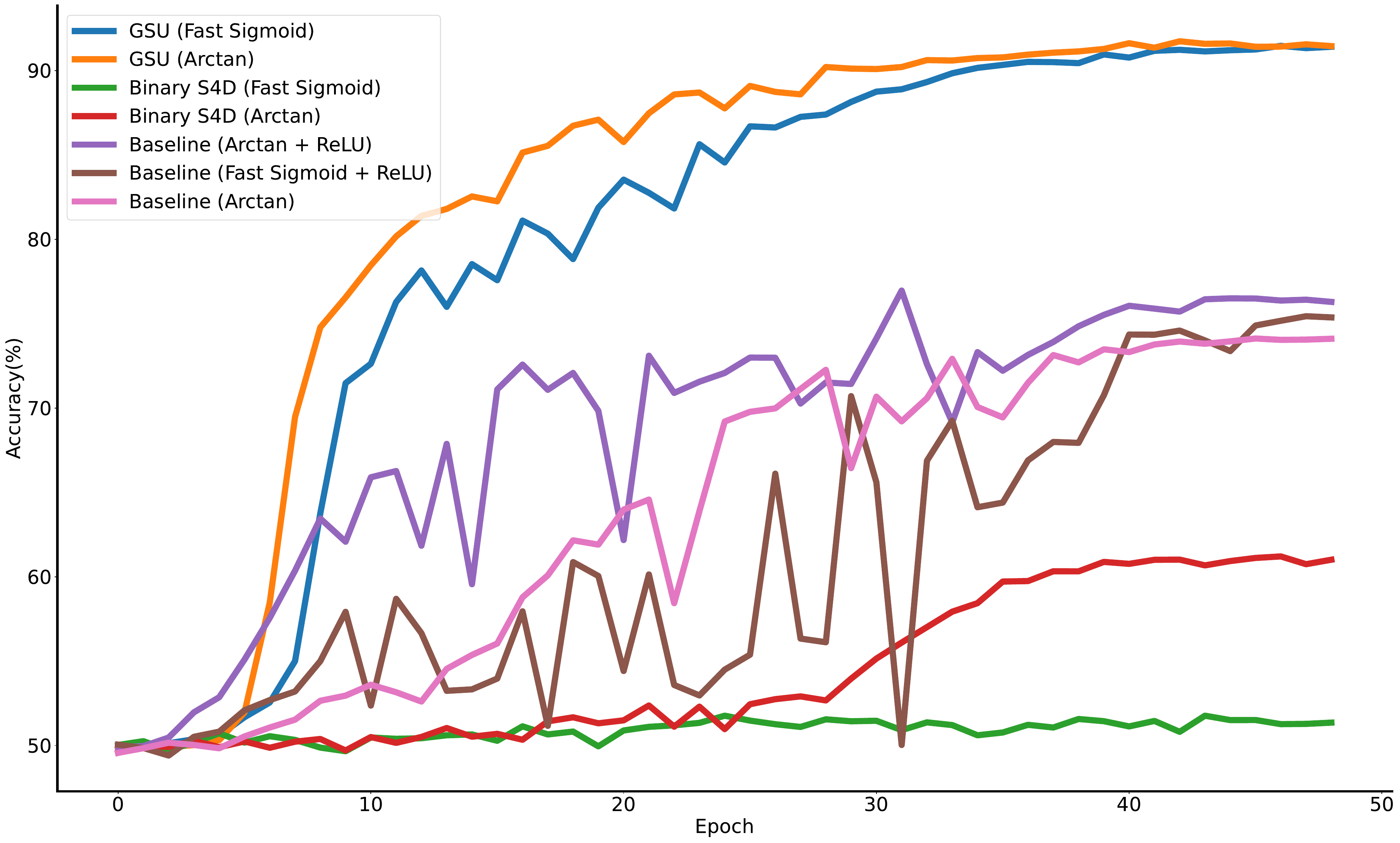}
        \hfill
        \caption{\textbf{Convergence on Path-X.} Applying saturating activation functions to SSM outputs leads to reduced accuracy on \textbf{Path-X}, similar to binary spiking activations.}
        \label{fig:pathx_performance}
\end{figure}

\begin{table}[H]
    \centering
    \begin{tabular}{lllllll}
        \toprule
        Model                        & No. Layers & Hidden Layer Size  & Surrogate Gradient Function & Accuracy \\
        
        \midrule
        
        \multirow{3}{*}{Binary S4D}  & 4                & 128          & Fast Sigmoid               & 69.62\%  \\
                                     & 4                & 128          & Arctan                     & 79.33\%  \\
                                     & 6                & 512          & Fast Sigmoid               & 69.83\%  \\
                                     & 6                & 512          & Arctan                     & 82.00\%  \\
                                     
        \midrule
        
        \multirow{3}{*}{GSU}         & 4                & 128          & Fast Sigmoid               & 80.11\%  \\
                                     & 4                & 128          & Arctan                     & 82.49\%  \\
                                     & 6                & 512          & Fast Sigmoid               & 85.01\%   \\
                                     & 6                & 512          & Arctan                     & 85.0\%   \\

        \bottomrule
        
    \end{tabular}
    \caption{Accuracy on \textbf{sCIFAR10} (\textbf{Image} from \textbf{LRA})}
    \label{cifar_table}
\end{table}

\subsection{\textbf{Effect of Surrogate Gradient Function}} \label{sec:surr_grad}

Two different functions are evaluated to highlight how sensitive Binary S4D and the GSU are to surrogate gradient choice. Previous research suggests that the choice of surrogate gradient function can impact the accuracy of an SNN, with arctan surrogates generally preferred over others \citep{eshraghian2023training}. Table~\ref{cifar_table} and Figure~\ref{fig:pathx_performance}, reinforce this observation. Binary S4D trained with arctan surrogate gradients achieves 79.33\%  and 82.00\% on \textbf{sCIFAR10}, in the smaller and larger model configurations, respectively. As one could reasonably expect, increasing the model size improves accuracy. In contrast, when using fast sigmoid surrogate gradients, accuracy falls to 69.62\% for the smaller configuration and to 69.83\% for the larger one. Therefore, fast sigmoid surrogate gradients cause a more than 10\% drop in accuracy, below 70\%, regardless of the size of the Binary S4D model. The trends identified on \textbf{sCIFAR10} are also replicated when analysing the results on \textbf{Path-X} (Figure~\ref{fig:pathx_performance}). Binary S4D with arctan surrogate gradients reaches 61.2\% in accuracy, while fast sigmoid gradients cause the accuracy to collapse to near-random (51.7\%). Hence, Binary S4D is generally sensitive to the surrogate gradient function used. 

When analysing GSU results, the effect of the surrogate gradient choice is greatly diminished. For the smaller configuration on \textbf{sCIFAR10}, adopting arctan boosts accuracy by more than 2\%, from 80.11\% to 82.49\%, in the smaller network size configuration. For the larger configuration, accuracy is essentially unchanged between the two surrogate functions, both approximately equalling 85\% (Table~\ref{cifar_table}). For the GSU, training on \textbf{Path-X} is also nearly indistinguishable between the two surrogates, although convergence is slightly faster for arctan surrogates than fast sigmoid (orange and blue curves in Figure~\ref{fig:pathx_performance}). 

\subsection{\textbf{Baseline Saturating Activations}} \label{sec:baseline_saturating}

Results in Section~\ref{sec:surr_grad} suggest that the choice of surrogate gradient function can impact accuracy, especially for Binary S4D. The evaluation of continuous saturating activations is used in this section to explore the potential performance of Binary S4D on long sequences if, hypothetically, an optimal surrogate gradient method was developed. 

When replacing spiking activations in Binary S4D with baseline continuous saturating activations and keeping all other hyperparameters unchanged, accuracy improves to some extent on \textbf{Path-X}. The baseline \textbf{Arctan + ReLU} and \textbf{Fast Sigmoid + ReLU} networks achieve 76.4\% compared to 75.44\%, respectively (Figure \ref{fig:pathx_performance}). This contrasts the higher sensibility of Binary S4D to surrogate gradient function selection, where fast sigmoid surrogates failed to converge beyond random selection accuracy. In addition, the GSU outperforms all baseline networks with continuous saturating activations, reaching 91.6\%  and 91.4\% with arctan and fast sigmoid surrogate gradients. This suggests that the inherently saturating behaviour of binary spiking activations significantly influences the degraded accuracy. 

Both arctan and fast sigmoid functions have negative outputs for negative inputs and intersect the origin (Section~\ref{sec:method_surr_grad}). Nesting the saturating activations within ReLU, emulates the subthreshold regime of binary spiking activations, where the output would be zero for negative inputs. Consequently, this runs the risk of ``dead neurons'' (those which always output zero and thus cannot learn) \citep{eshraghian2022fine, douglas2018relu}, which may affect model performance. Nonetheless, the baseline \textbf{Arctan} trained model manages to reach 74.12\% on \textbf{Path-X}, slightly lower than  \textbf{Arctan + ReLU}  (pink and purple curves in Figure~\ref{fig:pathx_performance}). This could mean that including ReLU does not produce ``dead neurons''  to a degree that would damage accuracy, and by extension, binary spiking SSMs may not be heavily affected by this phenomenon either. 

In answering question \textbf{(a)} from Section~\ref{sec:intro}, the results here suggest that binarisation does not render SSMs completely uncompetitive since they can still outperform the Transformer (Section~\ref{sec:lra_acc}) and state-of-the-art SNNs (Section~\ref{sec:smnist_acc}). However, this section provides evidence that it does inherently lower their performance. Regardless of the surrogate gradient function, binarisation is still a saturating activation. Hence, the upper bound of binary spiking accuracy is taken to be that of continuous saturating activations \citep{roberts2022principles}, which experiments in this section show to be lower than non-saturating counterparts such as the GSU, given all other factors are constant.

\section{Discussion}
This work has explored the effect of output binarisation in state-of-the-art SSMs in order to assess the viability of SSM-based SNNs as alternatives to ANN sequence models (question \textbf{(a)} in Section~\ref{sec:intro}). Results show that binarisation lowers accuracy to some extent compared to baselines, and the degradation is inherent to the saturating nature of binary spiking. Sections~\ref{sec:lra_acc} and \ref{sec:baseline_saturating} highlight how the GSU can overcome the vanishing gradient challenges of binary spikes while retaining efficient addition/subtraction-based feature mixing. This suggests that exclusively binary activations may not be necessary for SNNs, providing an answer to question \textbf{(b)} from Section~\ref{sec:intro}. 

Section~\ref{sec:smnist_acc} helps compare Binary SSMs with traditional SNNs. The reset mechanism in LIF neurons may help keep membrane voltages (preactivations) more closely centred around the threshold when firing. This means that when spikes occur, the gradient with respect to the membrane voltage is more likely to be from the non-saturate regime of the surrogate derivative \citep{herranz2022surrogate}, helping mitigate vanishing gradients. In contrast, Binary SSMs lack resetting mechanisms, meaning preactivations may stray further from the firing threshold. Arguably, this may cause binary spiking in SSMs to be more strongly affected by vanishing gradients than in LIF neurons. SSM normalisation strategies could potentially be employed in future work to help avoid this pitfall \citep{orvieto2023resurrecting}. Nevertheless, Section~\ref{sec:smnist_acc} shows how the SSM backbone can still help Binary S4D outperform current state-of-the-art SNNs on \textbf{sMNIST}, with fewer parameters.

Given the results in Section~\ref{sec:surr_grad}, one can infer that the choice of surrogate gradient determines, to a certain extent, the accuracy of the Binary SSM. This is best highlighted by the complete failure of Binary S4D to converge on \textbf{Path-X} when using fast sigmoid surrogate gradients, compared to the 61.1\% accuracy when using arctan (Figure~\ref{fig:pathx_performance}). Furthermore, the discrepancy between training with surrogate gradients and equivalent continuous activations underlines that there is potential for improving surrogate gradient training.  However, the disparity between baseline continuous saturating activations and the GSU (Section~\ref{sec:baseline_saturating}) highlights the intrinsic limitation that binary spiking activations inherit from saturating counterparts - vanishing gradients \citep{gulcehre2016noisy}.

Non-saturating activations such as ReLU are known to allow the construction of much deeper models than saturating nonlinearities \citep{glorot2011deep}, effectively avoiding vanishing gradients. The results in Section~\ref{sec:lra_acc} reflect this fact. The GSU, which allows the propagation of non-saturating values via ternary weights, outperforms Binary S4D on all tasks of the \textbf{LRA}. Section~\ref{sec:baseline_saturating}, shows how continuous saturating activation functions are also outperformed by the GSU on \textbf{Path-X}. In addition, the GSU manages to incorporate spiking nonlinearities while retaining comparable accuracy to the baseline S4D model on \textbf{LRA} (Section~\ref{sec:lra_acc}). These results suggest that the saturating behaviour of binary spiking, rather than its discontinuity, limits SNN performance. The overarching observation is that while there is room for improving surrogate-gradient training for SSM-based SNNs, even an ideal unbiased surrogate would struggle to compete with non-saturating activations. Hence, one could argue that implementing state-of-the-art large-scale SSM architectures on neuromorphic hardware should also include efficient forward propagation of non-saturated values. The proposed GSU shows that this is possible while still only using efficient addition/subtraction-based feature mixing. 

Certain neuromorphic platforms, such as Intel's Loihi, have begun to support integer graded spikes \citep{davies2021advancing, orchard2021efficient}. Therefore, the feasibility of techniques such as the proposed GSU could hinge on developing quantisation and sparsification strategies for the weights and activations of this new class of SNNs. One could also argue that current SNN methodologies already make use of propagating integer values. For example, effective residual connections, employed by state-of-the-art SNNs such as SpikeGPT, rely on spike-addition \citep{fang2021deep}. This results in layerwise integer outputs that scale with network depth, which differ from binary spikes \citep{chen2023training}. 

The contributions of this paper can be summarised as follows. First, this study formulates SSM-based SNNs and tests SNNs for the first time on the \textbf{LRA}, which contains sequence learning tasks with lengths much larger than traditional benchmarks used in neuromorphic research \citep{eshraghian2023training}. Moreover, for the first time, it is shown that SNNs can outperform Transformers on these established long-range sequence benchmarks. Second, this work demonstrates that SSNs built using SSMs can outperform current state-of-the-art SNNs on \textbf{Sequential MNIST}, while using fewer parameters. Finally, this work provides evidence to suggest that the saturating behaviour of spiking activations, not necessarily their discontinuity, can be considered the main challenge to scaling SNNs for long sequences and larger models. By introducing the GSU, it is further highlighted how this problem can be avoided without using dense vector-matrix multiplications relying on MAC operations. 

The significance of this paper's contributions stems from working towards bringing powerful SSMs to energy-efficient neuromorphic hardware. Recently proposed large language models based on SSMs have shown great potential in rivalling and even outperforming Transformer-based architectures \citep{dao2022hungry, poli2023hyena, gu2023mamba}, all while avoiding quadratic computational costs. Binary S4D and the GSU retain to a great extent the desirable properties of SSMs for sequence modelling, as highlighted by outperforming the Transformer on the \textbf{LRA}. This paves the way for deploying SSM-based SNNs to neuromorphic hardware, which could drastically reduce the energy requirements of sequential models. Taking into consideration the efficient scaling of computations with respect to sequence length, SSM-based SNNs could have the potential to replace current solutions such as GPU-deployed GPT4 \citep{openai2023gpt}. 

\section{Methods}

\subsection{\textbf{Leaky Integrate-and-Fire Neurons}} \label{sec:lif}

Spiking networks are most commonly built using Leaky Integrate-and-Fire (LIF) neurons \citep{eshraghian2023training}. They consist of discretising a simplified RC circuit dynamical system (Equation \ref{eq:lif}) \citep{gerstner2014neuronal}. Input currents ($i_t \in \mathbb{R}$) are linearly accumulated within the membrane voltage ($u_t \in \mathbb{R}$) of the neuron (Equation \ref{eq:disc_lif}). Current leakage refers to the exponential decay of inputs over time, controlled by the time constant $\tau \in \mathbb{R}$ and its discrete-time equivalent ($\beta \in \mathbb{R}$) (Equation \ref{eq:disc_tau}). Once the membrane potential crosses the firing threshold ($\theta \in \mathbb{R}$), a spike $s$ is emitted (Equation \ref{eq:spike}). As spikes are discrete events highly localised in time, they can be represented by either presence or absence, i.e. binary values $s \in \{ 0, 1\}$. Firing is followed by a refractory period when spiking is more difficult. This is implemented using feedback connections, whereby spiking causes the membrane voltage to be either set to a reset value or the threshold value $\theta$ is subtracted from the membrane potential (Equation \ref{eq:reset_connection}). This reset mechanism imposes iterative computations at training time, much like nonlinearities in RNNs (Figure~\ref{fig:unroll_rnn}). Removing feedback connections converts LIF neurons into LTI filters, which can be implemented as parallelisable convolutions. Equation~\ref{eq:lif_convolution}, shows this convolutional view in continuous time, with $\kappa$ being the global kernel implicitly parametrised by $\beta$. 

\begin{equation}
	\tau \frac{du(t)}{dt} = -u(t) + iR
\label{eq:lif}
\end{equation}

\begin{equation}
	u[t] = \beta u[t - 1] + (1 - \beta) i[t]
    \label{eq:disc_lif}
\end{equation}

\begin{equation}
    u[t] = \begin{cases}
        u[t], & s[t-1] = 0 \\
        u[t] - \theta, & s[t-1] = 1
    \end{cases}
    \label{eq:reset_connection}
\end{equation}

\begin{equation*}
    \beta = e^{\frac{-\Delta t}{\tau}}
    \label{eq:disc_tau}
\end{equation*}

\begin{equation}
    s[t] = \begin{cases}
                    1, & u[t] > \theta\\
                    0, & u[t] \leq \theta
                \end{cases}
    \label{eq:spike}
\end{equation}

\begin{equation}
    u(t)=\int_{0}^{\infty} \kappa(s)i(t-s)ds
    \label{eq:lif_convolution}
\end{equation}

\begin{equation*}
    \kappa(t) = \beta^{t}*(1 - \beta)
\end{equation*}

\subsection{\textbf{State Space Models}} \label{sec:ssm}

State space models (SSMs) are widely used tools in fields such as engineering and neuroscience \citep{gu2021efficiently}. Borrowing LIF concepts, SSMs can be understood as first projecting one-dimensional input currents ($i(t) \in \mathbb{R}$) onto higher dimensions, using a vector $B \in \mathbb{R}^d$, and the result is added to the state of the model ($u(t) \in \mathbb{C}^d$) (Equation \ref{eq:ssm}). The state is then propagated forward in time using a transition matrix $A$. For the purposes of this work, $A \in \mathbb{C}^d$ is taken to be diagonal. Complex values are required by $A$ and also passed on to the state $u$ in order to retain the same level of expressivity as most full-rank $d$x$d$ real matrices \citep{gu2022parameterization}. The state $u_t$ is then projected back to scalar output values ($y_t \in \mathbb{R}$) using $C \in \mathbb{R}^d$ and taking the real part of the product (Equation \ref{eq:ssm}). SSM parameters $A$ and $B$ in architectures such as S4 \citep{gu2021efficiently} are typically parameterised in continuous time, and a discretisation scheme is required. All experiments reported in this work are conducted using bilinear discretisation, following \citet{gu2022parameterization, gu2021efficiently} (Equation \ref{eq:bilinear_disc}). The time step size parameter $\Delta$ in Equation \ref{eq:bilinear_disc} also plays the role of determining how quickly the kernel decays over time, setting its time scale \citep{gu2020hippo, gu2022parameterization}. $I$ in Equation \ref{eq:bilinear_disc}, represents the $d \times d$ identity matrix. At training time, the discretised parameters $\overline{A}$ and $\overline{B}$, along with $C$, can be used to precompute the global convolution kernel ($\overline{K}$) (Equation \ref{eq:ssm_convolution}) for each batch. The convolutional theorem (Equation \ref{eq:conv_theory}), states that element-wise multiplication ($\odot$) in the Fourier domain is equivalent to convolution in the time domain. This means the computational cost is dominated by the Fourier transformation ($\mathcal{F}(.)$), and its inverse ($\mathcal{F}(.)^{-1}$), which can be computed efficiently in discrete settings using Fast Fourier Transforms (FFTs) in $\mathcal{O}(Llog{}(L))$ time, for sequence length $L$.

\begin{equation}
    \begin{split}
        u'(t) & = Au(t) + Bi(t) \\
        y(t) & = Cu(t) + Di(t)
    \end{split} 
    \label{eq:ssm}
\end{equation}

\begin{equation}
    \begin{split}
        \overline{A} & = (I - \Delta/2A)^{-1}(I + \Delta/2A) \\
        \overline{B} & = (I - \Delta/2A)^{-1} \cdot \Delta B
    \end{split}
    \label{eq:bilinear_disc}
\end{equation}

\begin{equation}
    \begin{split}
        y[t] & = \Sigma_{p = 0}^{t} \overline{K}[p] \cdot i[t - p] \\
        \overline{K}[p] & = C \overline{A}^{p} \overline{B}
    \end{split}
    \label{eq:ssm_convolution}
\end{equation}

\begin{equation}
    \overline{K} * i = \mathcal{F}^{-1}(\mathcal{F}(\overline{K}) \odot \mathcal{F}(i)) 
    \label{eq:conv_theory}
\end{equation}
    
\subsection{\textbf{State Space Initialisation}} \label{sec:initialisation}

SSM memory properties are deeply influenced by the choice of initialisation for $A$ and $B$. The eigenvalues of $A$ determine the asymptotic behaviour of $A^p$ required in computing $\overline{K}$ (Equation \ref{eq:ssm_convolution}), as $p \rightarrow \infty$. Since $A$ is taken to be diagonal, the eigenvalues ($\lambda_n$) are just its entries ($A_n$), which have been established above as being complex. SSMs parametrise the real and imaginary parts of these eigenvalues to encode an orthogonal basis. To ensure long-term stability and avoid exponential growth for large $p$, the real parts need to be negative ($Re(A_n) < 0$). \citet{gu2022parameterization} present $-\nicefrac{1}{2}$ to be an optimal choice for initialising the real part. During training, to ensure that the real part remains negative, it is typically enclosed within an exponential function ($ -e^{ln(Re(A_n))}$) \citep{orvieto2023resurrecting, gu2022parameterization}. The imaginary parts of the eigenvalues $Im(A_n)$ determine the spectral distribution of the basis and thus the expressivity of the global kernels $\overline{K}$ they span. To avoid confusion with the input ($i$), the imaginary unit in Equations~\ref{eq:S4D-Lin} and \ref{eq:S4D-Inv} is denoted by $j$. \citet{gu2022parameterization} propose several initialisation strategies for $Im(A_n)$, for example, \textbf{S4D-Lin} employs linearly-spaced $Im(A_n)$ (Equation \ref{eq:S4D-Lin}). S4D-Lin implements a damped Fourier basis, which, notably, has been examined in neuromorphic research before, e.g. within Resonate-and-Fire neurons \citep{orchard2021efficient} and resonator reservoirs \citep{hermans2010memory}. \textbf{S4D-Inv} is another proposed initialisation scheme, where $Im(A_n)$ are distributed by an inverse law (Equation \ref{eq:S4D-Inv}). S4D-Inv has been shown to outperform \textbf{S4D-Lin} on the \textbf{LRA}, especially \textbf{Path-X} \citep{gu2022parameterization}, therefore all models in this work are based on the S4D-Inv initialisation scheme.  

\begin{equation}
    A_{n} = -\frac{1}{2} + j\pi n
    \label{eq:S4D-Lin}
\end{equation}

\begin{equation}
    A_{n} = - \frac{1}{2} + j\frac{d}{\pi}(\frac{d}{2n + 1} - 1)
    \label{eq:S4D-Inv}
\end{equation}

\subsection{\textbf{Binary S4D}} \label{sec:binary_s4d}

As highlighted in Figure \ref{fig:binary_s4d}, the Binary SSM models examined in this work are built by applying the spiking function from LIF neurons, without the reset, to the scalar outputs ($y[t]$) of each independent SSM (Equation \ref{eq:binarisation_ssm}). In all experiments reported here, the firing threshold ($\theta$) is set to zero. The baseline SSMs being binarised are parametrised using the \textbf{S4D-Inv} scheme. Hence, Binary SSM models are referred to as Binary S4D throughout. The binary spikes ensure that feature mixing between different SSM channels does not require dense vector-matrix multiplication. However, it should be mentioned that some additional MAC operations are present in Binary S4D compared to LIF neurons. Namely, one can notice that integrating inputs in high-dimensional states $u$ is more expensive than scalar membrane voltages. Moreover, the dimensionality reduction step $Cu[t-1]$ in Equation \ref{eq:ssm} also requires additional MAC operations compared to LIF neurons. These added operations may increase energy costs over traditional LIF neurons. However, the focus here is on the effect on the accuracy of binary spiking activations. Energy-efficient neuromorphic implementations of Binary S4D can be reserved for future work. For example, \citet{voelker2019legendre} propose using population spike probability to implement SSM state operations at inference. 

\begin{equation}
    s(y[t]) =   \begin{cases}
                    1, & y[t] > \theta\\
                    0, & y[t] \leq \theta
                \end{cases}
\label{eq:binarisation_ssm}
\end{equation}

\subsection{\textbf{Surrogate Gradients}} \label{sec:method_surr_grad}

To account for the non-differentiability of the binary spike function, surrogate gradients are used in the backward pass of the training process \citep{neftci2019surrogate}. The gradients of two functions are adopted here - fast sigmoid (Equation \ref{eq:fast_sigmoid}) and arctangent (Equation \ref{eq:atan}). The hyperparameter $\alpha \in \mathbb{R}$ in Equation \ref{eq:fast_sigmoid}, is set to 25, following the defaults of the \textbf{snnTorch} library \citep{eshraghian2023training}. For the baseline saturating activations with continuous values tested on \textbf{Path-X}, each function is nested in a ReLU activation. As argued in Section~\ref{sec:baseline_saturating}, this is done to emulate the subthreshold behaviour of binary spiking activations. The arctan activation is also tested without ReLU nesting to check whether saturating at zero may result in ``dead neurons'' that negatively impact training. 

\begin{figure}[H]
	\centering
        \includegraphics[width=0.4\textwidth]{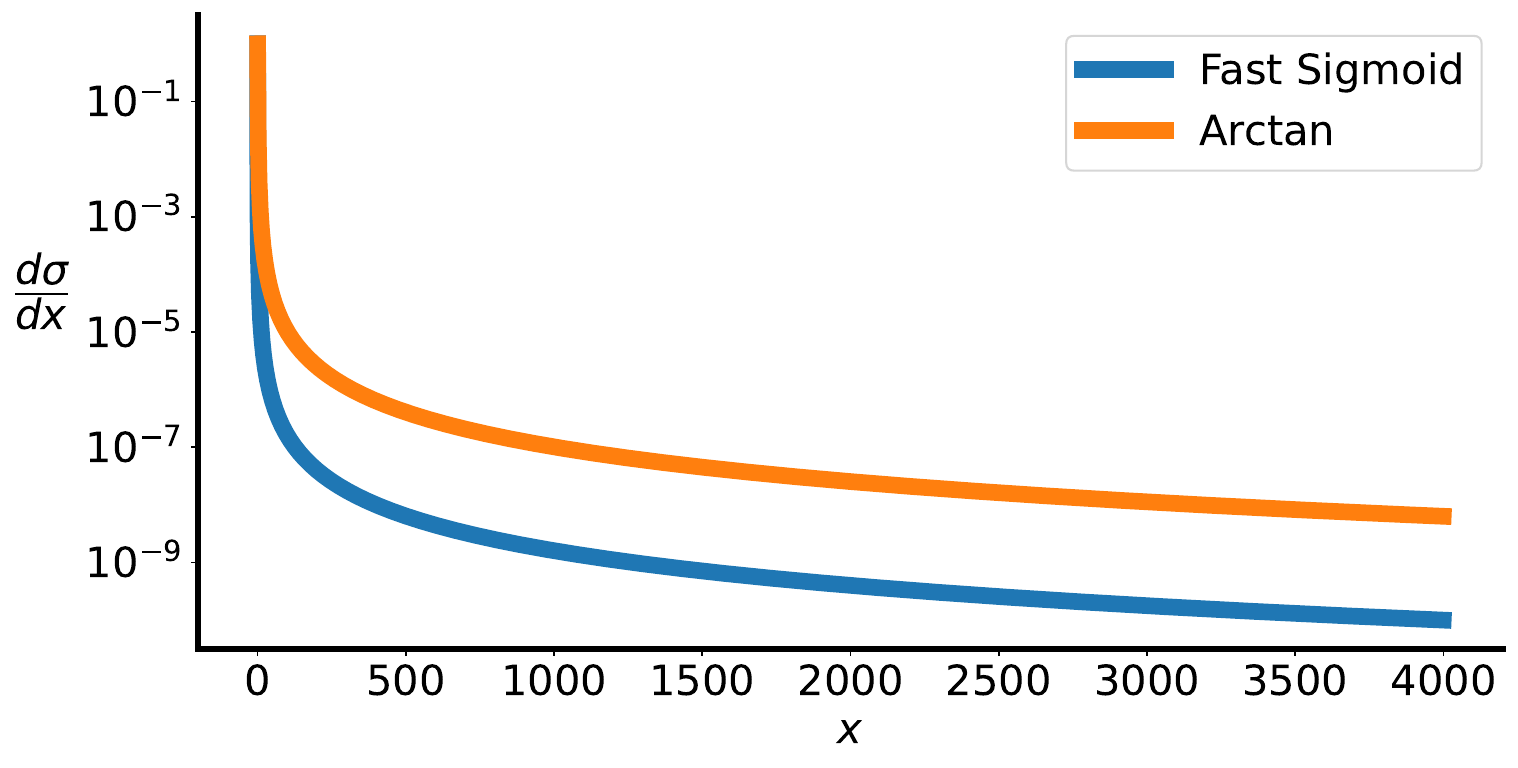}
	\caption{Fast sigmoid and arctan gradients decaying on a log-scale. Arctan gradients decay slower than fast sigmoid as $x \rightarrow \infty$. }
	\label{fig:gradients}
\end{figure}

\begin{equation}
    \begin{split}
        \sigma(x) & = \frac{x}{1 + abs(x) * \alpha} \\
        \frac{d\sigma}{dx} & = \frac{1}{(\alpha * abs(x) + 1)^2}
    \end{split}
    \label{eq:fast_sigmoid}
\end{equation}

\begin{equation}
    \begin{split}
        \sigma(x) & = \frac{1}{\pi} * arctan(\pi * x)\\
        \frac{d\sigma}{dx} & = \frac{1}{1 + (\pi x)^2}  
    \end{split}
    \label{eq:atan}
\end{equation}

\begin{equation}
    \sigma = ReLU(Fast Sigmoid(x)) \: or \:  ReLU(Arctan(x))
    \label{eq:relu}
\end{equation}

\subsection{\textbf{Gated Spiking Unit}} \label{sec:gsu}

Vanishing gradients arise in Binary S4D when constructing deep models since gradients may be greatly attenuated for early layers after passing through several saturating nonlinearities (binary spiking activations) \citep{gulcehre2016noisy}. This can be mitigated to some extent by using residual connections between layers \citep{fang2021deep, chen2023training}. However, even with residual connections, issues remain with gradient backpropagation to internal SSM parameters ($A, B, C, D$) since they can only flow through the saturating bottlenecks (binary spiking activations). The Gated Spiking Unit (GSU) is intended to serve as an example solution to avoid this bottleneck without introducing additional MAC operations. 

The GSU is inspired by the Gated Linear Unit (GLU) \citep{dauphin2016language}. GLU (Equation \ref{eq:glu}) passes inputs ($x \in \mathbb{R}^d$) through two linear projections in parallel, resulting in two feature vectors. A sigmoid nonlinearity ($\sigma$) is then applied to one of the vectors. The output of the GLU layer is the Hadamard product ($\odot$) of the two feature vectors, the sigmoid output acting as a scaling factor for the linear projection. Similarly, the GSU mixes input features ($x \in \mathbb{R}^d$) via two parallel routes (Equation \ref{eq:gsu}). First, each feature of $x$ is ternarised, i.e. continuous values $x_i$ are converted to values in $\{ -1, 0, 1\}$ following a thresholding method adapted from \citet{zhu2016trained} (Equation \ref{eq:ternary}). The parameter $\alpha \in \mathbb{R}$ is typically set to 0.15 by default and controls how sparse the nonzero features are. The ternary features $Ter(x)$ are then linearly projected using weights $W \in \mathbb{R}^{d \times k}$ and biases $b \in \mathbb{R}^{k}$. Simultaneously, $x \in \mathbb{R}^d$ is also multiplied by the ternarised weights $Ter(W) \in \{-1, 0, 1\} ^{d \times k}$ and biases $c \in \mathbb{R}^k$ are added (the weights $W$  are shared between the two streams). Ternarising $W$ works by computing the maximum function in $\Delta_W$ and iterating $W_{ij}$ over both dimensions of the weight matrix in Equation \ref{eq:ternary}. The output of the GSU is the Hadamard product of the two streams. One can observe that both matrix operations, $Ter(x) * W$ and $x * Ter(W)$, can be implemented using additions/subtractions, which are efficient mask operations \citep{yao2023spike}. It can also be noted that gradients can flow to both $x$ and $W$ via non-saturating routes, avoiding vanishing problems. In all experiments in this paper where it is present, the GSU layer is also followed by layer normalisation and Gaussian Error Linear Unit (GELU) activations \citep{hendrycks2016gaussian}. 

\begin{equation}
    GLU(x) = (x * W + b) \odot \sigma(x * V + c)
    \label{eq:glu}
\end{equation}

\begin{equation}
        GSU(x) = (Ter(x) * W + b) \odot (x * Ter(W) + c) 
\label{eq:gsu}
\end{equation}

\begin{equation}
    \begin{split}
        Ter(x_i) & = \begin{cases}
            x_i = 1,    &x_i >= \Delta_{x} \\
            x_i = -1,   &x_i <= -\Delta_{x} \\
            x_i = 0,    & otherwise \\
        \end{cases}\\
        \Delta_{x} & = \alpha * Max(Abs(x))
    \end{split}
    \label{eq:ternary}
\end{equation}

\begin{table}[H]
    \centering
    \resizebox{\textwidth}{!}{%
        \begin{tabular}{lllllllllll}
        \toprule
        Task       & No. Layers & No. Features  & Dropout & LR      & Batch Size    & Epochs    & WD    &  Norm & Pre-Norm  & ($\Delta t _{min}$, $\Delta t _{max}$)\\
        
        \midrule

        ListOps    & 8          & 128           & 0       & 0.01    & 50            & 40        & 0.05  &  BN   & False      & (0.001, 0.1)  \\        

        Text       & 6          & 256           & 0       & 0.01    & 16            & 32        & 0.05  &  BN   & True       & (0.001, 0.1)  \\ 

        Retrieval  & 6          & 256           & 0       & 0.01    & 32            & 11        & 0.05  &  BN   & True       & (0.001, 0.1)  \\  

        Image      & 6          & 512           & 0.1     & 0.01    & 50            & 200       & 0.05  &  LN   & False      & (0.001, 0.1)  \\  
        
        Pathfinder & 4          & 92            & 0       & 0.004   & 64            & 200       & 0.03  &  BN   & True       & (0.001, 0.1)  \\  

        Path-X     & 4          & 92            & 0       & 0.0005  & 32            & 50        & 0.05  &  BN   & True       & (0.0001, 0.1) \\  
        
        \bottomrule
        
    \end{tabular}}
    \caption{\textbf{LRA Experimental Configuration} WD refers to weight decay and LR to learning rate. BN signifies batch normalisation and LN layer normalisation.}
    \label{table:lra_hyperparameters}
\end{table}

\begin{table}[H]
    \centering
    \resizebox{\textwidth}{!}{%
        \begin{tabular}{llllll}
        \toprule
        Configuration                           &   Activation Function       & No. Layers        & No. Features  &  \textbf{Pathfinder} Acc. & \textbf{Path-X} Acc.\\
        
        \midrule
        
        Small (This Work)                       &  GELU                       & 4                 & 92            &  91.7\%            & 92.5\%    \\  

        Large \citep{gu2022parameterization}    &  GELU                       & 6                 & 256           & 93.78\%            & 92.80\%   \\  
        
        \bottomrule
        
    \end{tabular}}
    \caption{\textbf{Baseline S4D Accuracy on Pathfinder and Path-X} Because of the memory constraints of training on a single Nvidia A100 GPU, the sizes of the Binary S4D and GSU models used for \textbf{Pathfinder} and \textbf{Path-X} had to be reduced from the ones used by \citet{gu2022parameterization}. To provide a more accurate comparison in Sections~\ref{sec:lra_acc} and \ref{sec:baseline_saturating}, smaller baseline models are evaluated here. This table highlights how the baseline S4D with GELU activations used in this paper compare to the larger models employed by \cite{gu2022parameterization}. Besides the number of layers and features per layer, all other hyperparameters for the large models used by \cite{gu2022parameterization} are identical to the ones used here ( Table~\ref{table:lra_hyperparameters})}
    \label{table:baseline_s4d_param}
\end{table}

\subsection{\textbf{LRA Experimental Setup}} \label{sec:lra_config}

Overall, evaluation of the proposed methods on the \textbf{LRA} benchmark closely followed the experiments conducted in \citet{gu2022parameterization} to ensure that the object of the investigation is only the binary spiking activation (Table \ref{table:lra_hyperparameters}). More precisely, Binary S4D and the GSU are evaluated on \textbf{ListOps}, \textbf{Text} and \textbf{Image} using identical hyperparameters to \citet{gu2022parameterization}. \textbf{Retrieval} is evaluated using batch sizes reduced from 64 to 32 and fewer epochs (eleven compared to the original twenty). This is done to reduce training time and accommodate memory constraints on a single Nvidia A100 GPU. In addition, \textbf{Pathfinder} and \textbf{Path-X} are evaluated using smaller models than employed by \citet{gu2022parameterization} for the same reason. The baseline S4D-Inv results for \textbf{Pathfinder} and \textbf{Path-X} (Figure \ref{fig:acc_lra}) replicated here are obtained by replacing the binary spiking activation in Binary S4D with GELU activations and GLU feature mixing. In both \textbf{Pathfinder} and \textbf{Path-X}, \citet{gu2022parameterization} use six layers with 256 features each, compared to four layers with 92 features used here, reporting accuracies of 93.78\% and 92.80\%, respectively (Table~\ref{table:baseline_s4d_param}). Higher accuracies for the baseline S4D-Inv in the smaller configuration might have been possible if further hyperparameter tuning were employed. However, the goal here is only to isolate the effect of including binary spiking activation, all other hyperparameters being equal, such as the number of layers, SSM state size ($u \in \mathbb{C}^d$), number of epochs, etc. 

To maintain comparability with the S4D-Inv baselines, Binary S4D models use GLU layers for position-wise feature mixing after applying binary spiking activations. In addition, both Binary S4D and GSU networks are bidirectional. Finally, all models are implemented using addition-based residual connections. For Binary S4D, to ensure that only binary values are passed to the feature mixing layers, the residual addition takes place after feature mixing. The results in Figure~\ref{fig:acc_lra}, are obtained using arctan surrogates for both Binary S4D and the GSU in all tasks except \textbf{Retrieval}, where fast sigmoid surrogates are used. 

\subsection{\textbf{Sequential MNIST Experimental Configuration}} \label{sec:smnist_config}

The choice of hyperparameters for \textbf{sMNIST} classification is largely motivated by the intent to closely emulate traditional SNN computational principles. Hence, residual connections are removed in favour of exclusively spike-based communication between recurrent layers. Bidirectionality is also disabled. Models with both state dimensions ($dim(u) \in \{2, 64\}$) for GSU and Binary S4D are implemented in networks with two layers with 128 features (independent SSMs). 

\subsection{\textbf{Decoding}} \label{sec:decoding}

Temporal features from the last spiking SSM layer need to be compressed before being processed by the label prediction output layer for all of the classification tasks presented. The output of the final SSM layer can be viewed as a tensor $x$ of shape $[B \times L \times H ]$, where $B$ is the batch size, $L$ input sequence length, and $H$ is the number of hidden features. The output layer requires condensed inputs of shape $[B \times H]$. To reduce the $L$ dimension, average pooling is applied over it (e.g., \code{torch.mean(x, dim=1)} in PyTorch). This effectively entails that the final spiking layer is employing rate-coding. 

\section{Acknowledgements}
This research is supported in part through the NimbleAI project, which has received funding from the EU’s Horizon Europe Research and Innovation programme (Grant Agreement 101070679), and by the UK Research and Innovation (UKRI) under the UK government’s Horizon Europe funding guarantee (Grant Agreement 10039070).
See: \url{https://www.nimbleai.eu}.

\bibliographystyle{unsrtnat}
\bibliography{paper}  %%% Uncomment this line and comment out the ``thebibliography'' section below to use the external .bib file (using bibtex) .

\end{document}